# Neural Diffeomorphic–Neural Operator for Residual Stress–Induced Deformation Prediction

Changqing Liu[1], Kaining Dai[1], Zhiwei Zhao[1,2], Tianyi Wu[1], Yingguang Li[1,*]

1: College of Mechanical and Electrical Engineering, Nanjing University of Aeronautics and Astronautics, Nanjing, 210016, China

2: School of mechanical and aerospace engineering, Queen's University Belfast, Belfast, BT9 5AH, United Kingdom

*: Corresponding author: liyingguang@nuaa.edu.cn

## Abstract

Accurate prediction of machining deformation in structural components is essential for ensuring dimensional precision and reliability. Such deformation often originates from residual stress fields, whose distribution and influence vary significantly with geometric complexity. Conventional numerical methods for modeling the coupling between residual stresses and deformation are computationally expensive, particularly when diverse geometries are considered. Neural operators have recently emerged as a powerful paradigm for efficiently solving partial differential equations, offering notable advantages in accelerating residual stress–deformation analysis. However, their direct application across changing geometric domains faces theoretical and practical limitations. To address this challenge, a novel framework based on diffeomorphic embedding neural operators named neural diffeomorphic–neural operator (NDNO) is introduced. Complex three-dimensional geometries are explicitly mapped to a common reference domain through a diffeomorphic neural network constrained by smoothness and invertibility. The neural operator is then trained on this reference domain, enabling efficient learning of deformation fields induced by residual stresses. Once trained, both the diffeomorphic neural network and the neural operator demonstrate efficient prediction capabilities, allowing rapid adaptation to varying geometries. The proposed method thus provides an effective and computationally efficient solution for deformation prediction in structural components subject to varying geometries. The proposed method is validated to predict both main-direction and multi-direction deformation fields, achieving high accuracy and efficiency across parts with diverse geometries including component types, dimensions and features.

## 1. Introduction

Structural components are widely employed in advanced manufacturing to improve the performance of critical equipment, such as aircraft load-bearing beams, frames, and engine casings.[1] Accurate control of machining-induced deformation—both global and local—is essential not only for achieving dimensional precision but also for preserving functionality and structural integrity.[2] In practice, deformation control is typically realized through the optimization of machining processes aimed at regulating the deformation field. Such optimization requires thousands of deformation field predictions under varying process conditions to assess and compare their effectiveness in controlling deformation.[3] This challenge is further compounded by the large scale and geometric complexity of structural components, which place stringent demands on both the accuracy and efficiency of deformation prediction in order to enable effective control of the entire deformation field.

Deformation prediction involves estimating the deformation fields induced by different residual stress fields distributions within components, which corresponds to solving partial differential equations (PDEs) defined on complex and varying geometric domains.[4] Existing approaches for deformation prediction can be broadly categorized into numerical methods and data-driven methods[5]. Traditionally, numerical methods such as finite element analysis (FEA) are employed to solve the governing PDEs based on physical laws. Li et al.[6] developed a theoretical–numerical framework to predict machining distortion of pre-bent aircraft skin-panels by incorporating bending- and machining-induced stresses into a finite element sub-modeling approach. Vovk et al.[7] applied finite element simulations based on the Coupled Eulerian–Lagrangian method to analyze stress formation and residual stress effects on deformation in hard milling without the need for remeshing. While capable of delivering high accuracy, these methods are computationally expensive, particularly when the component size is large and when multiple geometries and loading conditions must be evaluated repeatedly during process optimization.

The data-driven methods have emerged as a promising approach for predicting machining deformation through leveraging their advantages to capture internal relationship among variables and achieve high computation efficiency. Sun et al.[8] proposed a hybrid-driven approach that integrates a Gaussian process–based surrogate model with a Bayesian framework to predict machining errors of thin-walled parts under uncertainties of cutting forces and tool wear. Zhao et al.[9] introduced a machining deformation prediction method that employs a physical-field estimation neural network to infer the latent residual stress field from deformation force data and predict subsequent deformation in large structural parts.

Among neural networks methods, physics-informed neural networks (PINNs) leverage learning to predict PDE solutions more efficiently and achieve higher accuracy embedding PDE constraints into network training, enabling surrogate modeling with fewer labeled examples and mesh-free evaluation. Wang et al.[10] proposed a neural architecture search–guided physics-informed neural network that automatically optimizes network architectures for solving PDEs through bi-level optimization with mask-based tensor operations. Zhang et al.[11] developed a convolutional physics-informed neural network with multiple receptive fields, high-order finite difference via Taylor boundary padding, and a dimensional balance method for loss weighting to improve accuracy and convergence in solving PDEs. However, pure data-driven models often struggle to generalize, and PINNs may suffer from optimization challenges and difficulty in handling complex geometries.

A more scalable paradigm for PDE solution learning is the neural operator (NO), which seeks to approximate mappings between infinite-dimensional function spaces, i.e., solution operators of PDEs. Once trained, neural operators can rapidly provide PDE solutions for new input functions, thereby offering a mesh-independent and efficient alternative to conventional solvers. Representative architectures include DeepONet[12], Fourier Neural Operator (FNO)[13], and graph-based neural operators[14], which have demonstrated success across diverse domains such as fluid dynamics, elasticity, and climate modeling. However, most existing neural operator frameworks assume a fixed computational domain. This limitation arises because functions defined on different geometries belong to different Banach spaces, making the operator mapping ill-posed when the domain varies. As a result, current NO methods are primarily restricted to problems with fixed domains. To address this issue, geometry-aware operator learning approaches have been proposed, such as the Geometric Operator Network (GINO), which extends operator learning to variable geometries and has achieved encouraging results[15]. A variation model of FNO named Geo-FNO was proposed to adapt to general geometries through incorporating geometry representation into NO model[16]. Hao et al. proposed a transformer-based NO framework named GNOT, which involves geometry by learning with heterogeneous normalized attention and geometric gating mechanisms for PDE solving on irregular meshes and multi-scale problems.[17] Nonetheless, such methods remain largely heuristic and lack a theoretical foundation, leaving geometry variability an open challenge in operator learning.

In the direction of operator learning over varying geometries, the Diffeomorphic Neural Operator (DNO) provides a theoretically grounded framework[18]. By introducing diffeomorphic mappings (e.g., harmonic mappings) to project complex domains into a common reference domain, DNO enables geometry-agnostic operator learning. Similar design was also proposed in DIMON, which calculates the mapping through an optimization process[19]. DIMON and DNO have provided theoretical guarantees and demonstrated effectiveness in applications such as fluid dynamics and solid mechanics. Empirical studies further show that DNO can generalize to previously unseen domains at significantly reduced computational cost. Beyond these approaches, latent diffeomorphic embeddings have also been explored to improve generalization, for instance by pulling solutions back to a canonical domain to reduce data requirements. Despite these advances, current diffeomorphic embedding methods still face notable limitations. Existing formulations are primarily restricted to two-dimensional geometries through harmonic mappings, which do not extend efficiently to three-dimensional domains. While alternatives such as Large Deformation Diffeomorphic Metric Mapping (LDDMM) can provide diffeomorphic mappings in 3D, they are computationally intensive and require case-specific optimization for each geometry, which severely limits their practicality[20]. This drawback is particularly critical for machining-induced deformation prediction of large and complex structural components, where many deformation field evaluations across varying geometries are required for process optimization.

To address this issue, this paper proposes a Neural Diffeomorphic-Neural Operator (NDNO) framework, that integrates a diffeomorphic mapping learning network into a neural operator model. This design transforms varying domain geometries into a reference domain to learn a unified solution operator for machining deformation prediction. Specifically, NDNO employs point clouds to represent complex 3D component geometries and constructs a registration NN to compute geometry mappings between different component geometries and a reference domain, driven by the constraints of geometric distance and diffeomorphism conditions. The registration NN facilities the model with efficient mapping computation in an end-to-end manner. Through a common reference domain, NDNO accommodates varying input geometries while preserving the physical relationships among variables in original domains through diffeomorphic mappings. Thereby, NDNO is capable of learning the solution operator of PDEs governing machining deformation across diverse component geometries, achieving high predictive accuracy and efficiency. The key innovation of this method lies in learning the mapping of varying geometries with the diffeomorphic neural network, which is driven and constrained by the loss function formulated by geometric similarity and mapping diffeomorphism. The overall structure of NDNO is illustrated in Fig. 1.

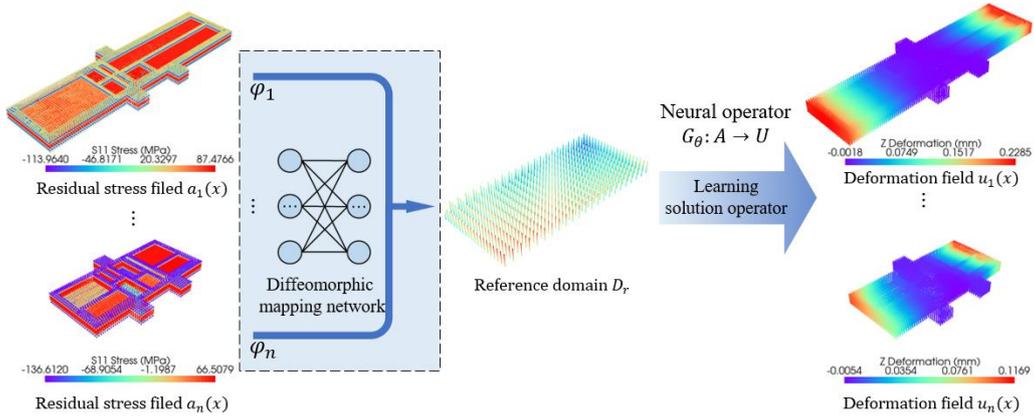

Fig. 1 Overall structure of NDNO

## 2. Problem Formulation

### 2.1. Equations for Deformation

Machining deformation in structural components is strongly influenced by residual stresses accumulated during upstream processes such as forging, quenching, or heat treatment[21]. Let $\Omega \subset \mathbf{R}^3$ denote the geometry of a structural component and let $\sigma(x)$ represent the residual stress field defined on $\Omega$. The machining process induces a redistribution of these stresses, leading to the development of a deformation field $u(x)$. The governing equations can be expressed in the general form of equilibrium and constitutive relations:

$$\nabla \cdot \left(\mathbb{C}: \nabla^s u(x) + \sigma(x)\right) = f(x), x \in \Omega, \tag{1}$$
$$u(x) = \bar{u}(x) \quad x \in \partial\Omega, \tag{2}$$

where $\nabla^s u(x)$ denotes the symmetric gradient, $\mathbb{C}$ is the fourth-order elasticity tensor, relating strain to elastic stress, $f(x)$ denotes body forces. $\bar{u}(x)$ denote prescribed displacements on the boundary $\partial\Omega$.

This system can be summarized as a nonlinear PDE operator $\mathcal{F}$, such that
$$\mathcal{F}(\sigma(x), u(x); \Omega) = 0, \tag{3}$$

Traditional numerical methods such as finite element analysis (FEA) provide accurate solutions to the above system but are computationally prohibitive when repeatedly applied to a family of components with varying geometries. In practice, structural components $\{\Omega_i\}_{i=1}^N$ differ significantly in shape, and solving the PDE for each geometry introduces severe efficiency bottlenecks.

*2.2. Diffeomorphic Neural Operator*

To address the limitations of conventional solvers, neural operators have been proposed as efficient surrogates to learn mappings between functional inputs and outputs governed by PDEs. Specifically, for machining deformation, the objective is to approximate the operator

$$\mathcal{G}_\theta: \sigma(x) \mapsto u(x), x \in \Omega, \tag{4}$$

where $\theta$ denotes trainable parameters. However, existing neural operator frameworks are typically trained on fixed domains and struggle to generalize to new geometric configurations. This limitation poses a fundamental obstacle for residual stress–induced deformation prediction across diverse structural components. The DNO framework was proposed to overcome this challenge, through integrating geometric embeddings with operator learning. Its key idea is to employ a diffeomorphic mapping,

$$\phi_\psi: \Omega_i \to \Omega_r, \tag{5}$$

where $\phi_\psi$ is a diffeomorphic mapping to map original domain $\Omega_i$ into a common reference domain $\Omega_r$. Residual stress and deformation fields defined on $\Omega_i$ are correspondingly transformed as

$$\tilde{\sigma}_i(x) = \sigma_i\left(\phi_\psi^{-1}(x)\right), \tilde{u}_i(x) = u_i\left(\phi_\psi^{-1}(x)\right), x \in \Omega_r. \tag{6}$$

On this unified reference domain, a neural operator $\mathcal{G}_\theta$ is trained to approximate the geometry-agnostic mapping
$$\tilde{u}(x) \approx \mathcal{G}_\theta(\tilde{\sigma}(x)), x \in \Omega_r. \tag{7}$$

Then, the overall framework is expressed as
$$u_i(x) \approx \phi_\psi^{-1}\left(\mathcal{G}_\theta(\sigma_i \circ \phi_\psi^{-1})(x)\right), x \in \Omega_i. \tag{8}$$

This construction enables accurate prediction of machining deformation across families of structural components, as the diffeomorphic embedding unifies heterogeneous geometries into a consistent learning domain.

## 3. Neural Diffeomorphic-Neural Operator

Earlier studies demonstrated the feasibility of DNO in two-dimensional domains using harmonic mappings. For three dimensions, existing alternatives such as LDDMM can provide diffeomorphic embeddings but are computationally inefficient, as they require optimization for each individual component. To address these challenges, we propose a new approach for efficiently constructing three-dimensional diffeomorphic embeddings.

In this study, a framework named NDNO is proposed aimed at learning the solution operator of PDEs governing machining deformation across varying component geometries. The proposed method represents the domain geometry and data in the form of point cloud to accurately express complex 3D geometries and subtle differences. A registration neural network is constructed to learn the diffeomorphic mapping that transforms varying domains into a common domain as a mapping sub-network. Embedding the mapping network, the neural operator is capable of learning the solution operator of the PDE governing a class of problem with varying geometries and efficiently handling unseen domain geometries.

This section provides a detailed description of the proposed method. Section 3.1 describes the structure of the mapping neural network. Section 3.2 details the constraint of the mapping network that drives the network to implement a diffeomorphic mapping. Section 3.3 outlines the overall framework of the proposed neural operator embedding the mapping sub-network.

*3.1. Diffeomorphic neural network*

The implementation of solving a class of PDEs defined in varying domains through learning the common solution operator requires mapping the domains into a reference domain. While the mapping computation of diverse geometries remains a challenge for the overall efficiency, especially with complex 3D geometries. Moreover, maintaining the physical relationship among variables in the PDEs during the mapping process also brings up extra threshold.

Therefore, a registration neural network is constructed to compute the corresponding mapping for each point cloud geometry in an end-to-end manner. The registration network takes the point cloud of the original geometry and the target geometry, i.e. the reference domain as its input, while also incorporating the target geometry into the governing loss function as a constraint. The network updates an incremental displacement field which transforms the input geometry to approach the target geometry in the form of point cloud. The structure of the mapping network is illustrated in Fig. 2.

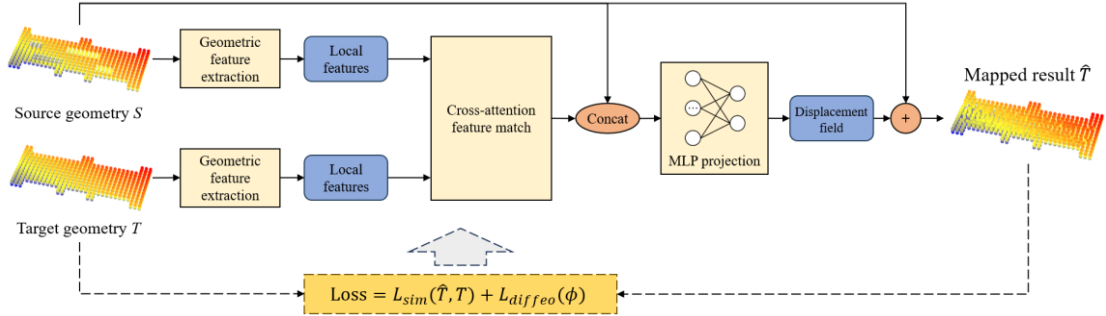

Fig. 2 Structure of the diffeomorphic neural network

The network consists of a feature extracting module, a feature matching module and a geometry morphing module. Dynamic graph convolutional neural network (DGCNN) is used to extract the local geometric feature of the source geometry and target separately. Then the extracted features match with each other through the cross-attention in a transformer block. Restored to the 3D space by multilayer perceptron (MLP) as the projection module, the matched features in high dimension are output as the displacement of each data point in the source point cloud.

The mapping network is further constrained for learning the displacement field to approach a diffeomorphic mapping towards the target geometry, implemented in the form of loss function. The constraint can be described as the target functional $H$ of the mapping network as:

$$H[\phi] = \min R(\phi) + \lambda \cdot D(\phi \circ S, T) \tag{9}$$

where $\phi$ is the mapping that the registration network implemented on the input geometry, $R$ is the diffeomorphism measuring function, $D$ is the distance function measuring the similarity between two geometries, $S$ is the source geometry, $T$ is the target geometry, and $\lambda$ is the coefficient of the distance term.

The specific property of the mapping related to diffeomorphism can be concluded as smoothness, invertibility and topology preservation. Invertibility and topology preservation of the mapping can be evaluated with Jacobian determinant. The Jacobian determinant represents the local volume scaling factor, used to describe the local geometric change during mapping process. The Jacobian determinant at a point $x_0$ in 3D space is defined as:

$$\det\bigl(J_\varphi(x_0)\bigr) = \begin{vmatrix} \frac{\partial \phi_x}{\partial x} & \frac{\partial \phi_x}{\partial y} & \frac{\partial \phi_x}{\partial z} \\ \frac{\partial \phi_y}{\partial x} & \frac{\partial \phi_y}{\partial y} & \frac{\partial \phi_y}{\partial z} \\ \frac{\partial \phi_z}{\partial x} & \frac{\partial \phi_z}{\partial y} & \frac{\partial \phi_z}{\partial z} \end{vmatrix} \in \mathbb{R}^{3\times 3} \tag{10}$$

where $\varphi_x$、$\varphi_y$、$\varphi_z$ denotes the components of the mapping in different directions.

The value of Jacobian determinant reveals how the local volume changes. When $\det(J_\varphi) < 1$, the deformation in the mapping causes compression in local geometry; $\det(J_\varphi) = 1$ indicates folding of the local structure. Folding results in multiple distinct points being mapped to the same location, thereby violating invertibility. Inversion occurs when the local normal vector of the surface is reversed, breaking topological consistency. Therefore, the Jacobian determinant is constrained to be positive by an inversibility loss term of the mapping network as:

$$L_{inv} = \frac{1}{N}\sum_{i=1}^{N} \mathrm{ReLU}\bigl(-\det(J(x_i))\bigr) \tag{11}$$

where $N$ is the point number, ReLU denotes the activation function that penalizes the non-positive Jacobian determinant value.

Smoothness of the mapping is evaluated with a Sobolev regularization term. The Sobolev regularization term is constructed by the squared L2 norm of the first order derivatives of the mapping function in the Sobolev space as:

$$\|\partial \varphi(x)\|^2 = \|\nabla \varphi\|^2 = \bigl\|J_\varphi(x)\bigr\|^2 \tag{12}$$

The Jacobian determinant is introduced to compute the gradient in the smoothness constraint, forming the smoothness loss term as:

$$L_{smooth} = \sum_{i=1}^{N} \|J(x_i) - I^3\|^2 \tag{13}$$

The Jacobian determinant of the mapping in the network is calculated through the automatic differentiation of neural network. Both of the invertibility and smoothness loss term forms the diffeomorphism term in the loss function:

$$L_{diffeo} = \beta_1 L_{rev} + \beta_2 L_{smooth} \tag{14}$$

where $\beta_1$ and $\beta_2$ are the coefficient of each loss term.

Moreover, the similarity between the source geometry and target geometry is calculated to drive the network transform the input geometry towards the target geometry. Since point cloud is composed of unorganized points and varying point numbers, the similarity between point cloud is evaluated as the difference between two distributions by Sinkhorn distance. Based on the optimal transfer that find a transfer with lowest cost, Sinkhorn distance incorporates entropy regularization to reinforce computation efficiency and robustness towards complex distributions[22]. The calculation of Sinkhorn distance in point cloud as discrete distribution is conducted with Sinkhorn-Knopp Algorithm. And the resulting similarity loss term is:

$$L_{sim} = \beta_3 \cdot d_{Sinkhorn}(\phi \circ S, T) \tag{15}$$

where $d_{Sinkhorn}$ denotes the Sinkhorn distance function, and $\beta_3$ is the coefficient of similarity loss term. The neural network updates the parameters with this similarity and drives the mapping transforms towards the geometry of the reference domain.

The overall loss function $L$ of the mapping network is represented as:

$$L = L_{diffeo} + L_{sim} = \beta_1 L_{rev} + \beta_2 L_{smooth} + \beta_3 \cdot d_{Sinkhorn}(\phi \circ S, T) \tag{16}$$

## 3.2. Neural operator architecture

The geometric information mapped to the reference domain by the mapping network is subsequently used in the NO. NO is employed in this study to learn the solution operator of the PDE across varying domains to predict component machining deformation. Defined in a finite dimension parameter space $\Theta$, the NO builds solution operator $G_\theta$ with network parameters $\theta$ to approximate the real solution operator $G^\dagger$ defined in infinite-dimensional space:

$$G^\dagger: A \to U \tag{17}$$

$$G_\theta: A \to U, \quad \theta \in \Theta \tag{18}$$

where $A$ and $U$ is separately the space that residual stress field and machining deformation field belongs to.

The learning target of the NO is formulated as updating the $\theta$ to reduce the discrepancy between $G^\dagger$ and learned $G_\theta$:

$$\min_{\theta \in \Theta} \left\| G^\dagger(\sigma) - G_\theta(\sigma) \right\|_U^2 \tag{19}$$

The NO in this study is formulated as an iterative structure that extracts the latent data features by layers, which is composed of linear part $W_t$, integral part $K_t$ and bias part $b_t$. Each layer is transferred to the next before a non-linear activation layer $\sigma_t$. Combined with the lifting layer $P$ and projection layer $Q$, the entire NO structure is represented as:

$$G_\theta \coloneqq Q \circ \sigma_T(W_T + K_T + b_T) \circ \ldots \circ \sigma_1(W_1 + K_1 + b_1) \circ P \tag{20}$$

This NO structure can be embedded with diverse specific NO layer designs, such as Laplace NO, Convolutional NO and Fourier NO. The framework in this study is constructed with the structure of Geo-FNO to enhance computation efficiency and integrate geometric information, composed of iterative Fourier layers that capture the global features in frequency domain. The overall structure of the NO is illustrated in Fig. 3. The NO model takes the input function of the PDE and the geometry information as the model input, and output the solution function at the query location. The location data is normalized before inputting for enhanced stability across varying geometries.

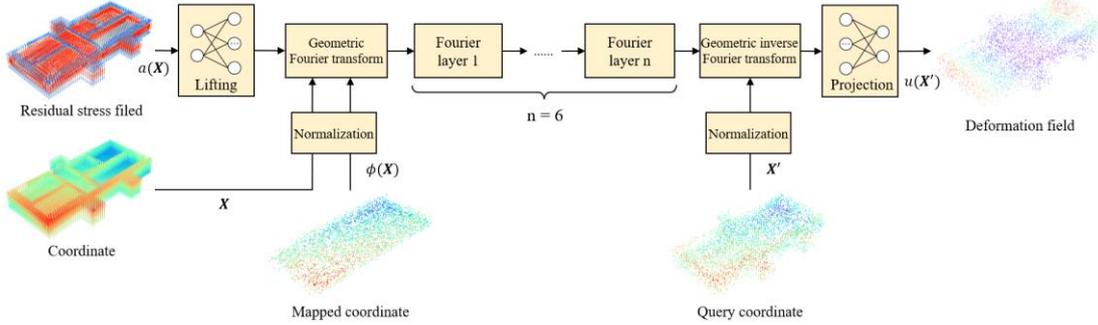

Fig. 3 Structure of the NO in NDNO

Since the data form of point cloud does not conform to the requirement of fast Fourier transform, the geometric information is also incorporated to transform the data into Fourier space through discrete Fourier transform and inverse Fourier transform. Moreover, the geometric information is also embedded as the NO input to reinforce the model adaptability to different geometries. The structure of the geometric Fourier transform module and inverse transform module is presented in Fig. 4.

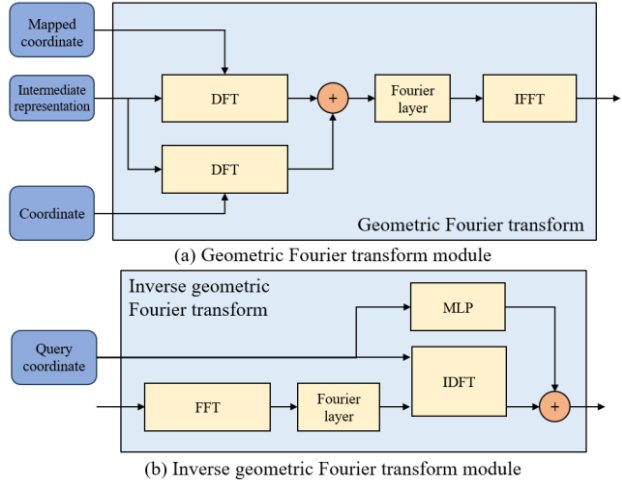

Fig. 4 Intermediate modules of the NO

Regarding the training process of the NO, the loss function is formulated as the relative error in L2 norm of the output solution function as:

$$Loss = \frac{\|\hat{y} - y\|_2}{\|y\|_2} \qquad (21)$$

where $\hat{y}$ is the predicted deformation field, $y$ is the label data of the deformation field.

Through the mapping network and the NO, the solution of complex relationship between variables in varying domains is implemented in the form of point cloud. And the solution function at query locations is predicted through learning the solution operator of PDEs.

## 4. Numerical Validation

This study was verified on prediction of machining deformation on diverse conditions including residual stress distributions and part geometries. The proposed framework takes the residual stress field and the component geometry as the input, and the deformation field as the output in the form of 3D point cloud. The residual stress field consists of stress in two directions; the deformation field consists of one main direction and two secondary directions. The method employed Adam optimizer to update model parameters and cosine annealing learning rate scheduler to reinforce convergence and stability in the training process. The main parameter of model training is listed in the Table 1.

Table 1 Main parameters in NDNO framework training

| Batch size | Learning rate | Epoch | $\beta_1$ | $\beta_2$ | $\beta_3$ |
|---|---|---|---|---|---|
| 4 | 1e-2 | 500 | 1e4 | 1e-3 | 1e5 |

### 4.1. Mapping network validation

The component geometry is used to validate the performance of the mapping network. A series of common registration network structures for point cloud are tested in comparison with the proposed cross-attention based diffeomorphic neural network, which were trained with identical loss function. The loss curve of the training process is illustrated in Fig. 5.

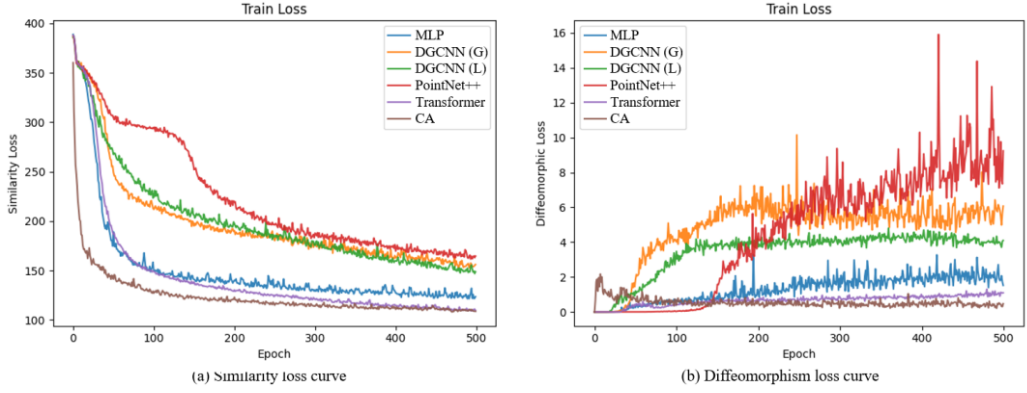

Fig. 5 Training loss curve of the mapping network

All the tested registration model is trained with identical loss in Eq. XX, constraining the diffeomorphism of the mapping. Moreover, the optimization-based registration model LDDMM is incorporated as a baseline for the relative similarity loss. The comparison result is measured by each loss terms averaged in the test dataset, as presented in Table 2.

Table 2 Loss terms of the diffeomorphic neural network test result

| Registration model | Similarity loss | Relative similarity loss | Diffeomorphism loss |
| --- | --- | --- | --- |
| MLP | 113.59 | 13.51 | 15.02 |
| DGCNN (Global) | 205.85 | 105.77 | 315.97 |
| DGCNN (Local) | 168.48 | 68.4 | 69.7 |
| PointNet++ | 178.21 | 78.13 | 88.28 |
| Transformer | 111.65 | 11.57 | 11.92 |
| Cross attention | **105.78** | **5.7** | **0.64** |

Both based on the transformer structure, the transformer and cross-attention registration model shows the best stability in the training process and similarity in the mapping result. Their performance is also similar to the result of optimization-based LDDMM. And the proposed cross-attention mapping network achieves the lowest diffeomorphism loss, which apparently outperforms other registration models. Consequently, the proposed model is capable of learning the geometry mapping that achieves high similarity and diffeomorphism.

The mapping result of tested models is also compared on the mapped geometries. The input geometries are components with machining features on the top surface, and are mapped to a blank with same dimension, as shown in Fig. 6. And the mapping result of the transformer and cross-attention mapping network is illustrated in Fig. 7 due to their performance out of all tested networks. It can be inferred that the proposed cross-attention network achieves apparently better result in similarity to the target geometry.

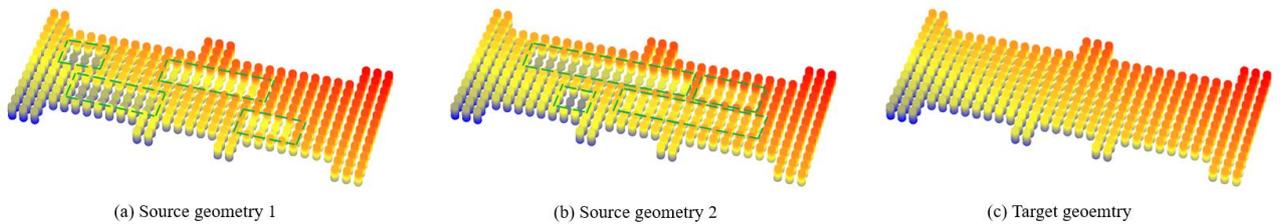

(a) Source geometry 1    (b) Source geometry 2    (c) Target geoemtry

Fig. 6 Examples of component geometries in diffeomorphic neural network

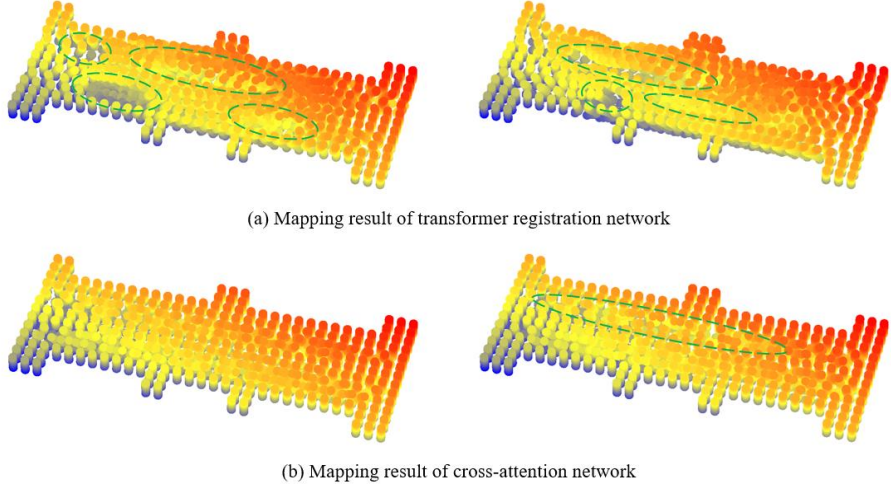

(a) Mapping result of transformer registration network

(b) Mapping result of cross-attention network

Fig. 7 Morphing result of the tested networks

In order to investigate the effect of the proposed constraining diffeomorphism loss on the mapping network, comparison models that is trained with only similarity loss is also tested. The mapping effect of tested models is visualized in displacement lines, as illustrated in Fig. 8. The displacement outputted by the similarity-based networks exhibits considerable crossing between the displacement path of points and long-distance displacement, which result in destruction in the geometry topology structure. And the long-distance displacement also exists in the result of the transformer registration network, though it improves in reducing crossing among displacement paths. As a comparison, the proposed cross-attention network achieved to approach the target geometry with a high proportion of local displacement, which was also distributed more uniformly in each direction. Therefore, it can be inferred that the diffeomorphism constraint apparently enhances the rationality and topology reservation of the mapping. And the mapping of the cross-attention network achieved higher uniformity in the displacement in three directions.

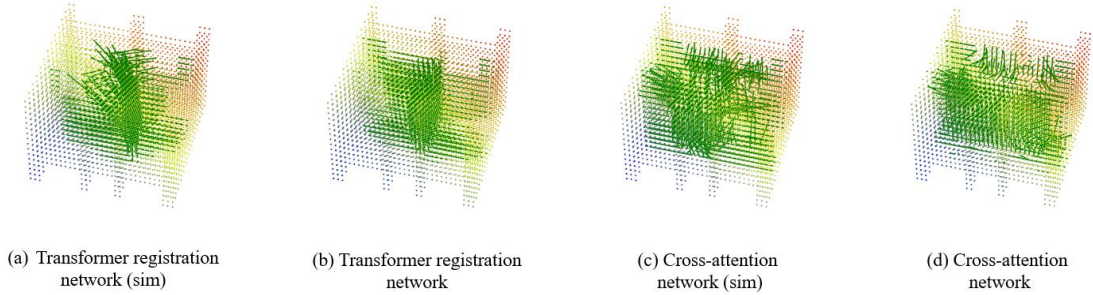

(a) Transformer registration network (sim)  (b) Transformer registration network  (c) Cross-attention network (sim)  (d) Cross-attention network

Fig. 8 Displacement field of the mapping network output

*4.2. Validation of predicted deformation*

The prediction result of the NO in DMEMO is validated on two different types of components, including frame and C-beam. The component geometry is presented in Fig. XX. Since the discrete Fourier transform mapping data to the frequency domain is set with basis function in a period, the coordinate of the input data and query location is initially normalized according to the median of dimension in each component type.

The prediction result on the main deformation direction on Z axis is tested on both component types, with examples shown in Fig. 9. The overall result of the predicted deformation is shown in Table 3 and Table 4. And the max error of all the tested samples is shown in the form of density distribution in Fig. 10.

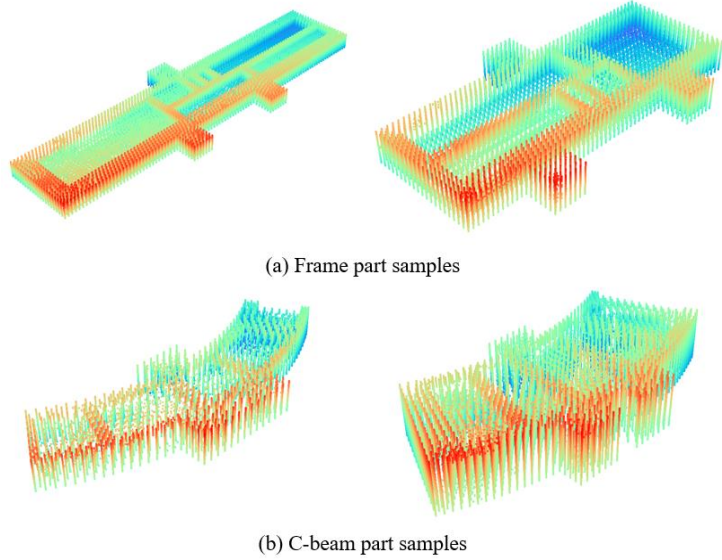

(a) Frame part samples

(b) C-beam part samples

Fig. 9 component geometry examples

Table 3 Main direction deformation prediction results on frame components

| - | NDNO | Geo-FNO |
|---|---|---|
| Averaged max error (mm) | **0.0452** | 0.0603 |
| RMSE (mm) | **0.0140** | 0.0188 |

Table 4 Main direction deformation prediction results on C-beam components

| - | NDNO | Geo-FNO |
|---|---|---|
| Averaged max error (mm) | **0.0273** | 0.0597 |
| RMSE (mm) | **0.0076** | 0.0090 |

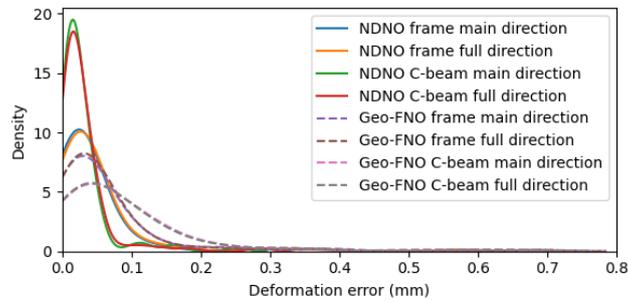

Fig. 10 Density distribution of deformation max error

The test result of the predicted machining deformation is also illustrated with the deformation and error distribution in examples in Fig. 11 and Fig. 12 to exhibit the prediction accuracy.

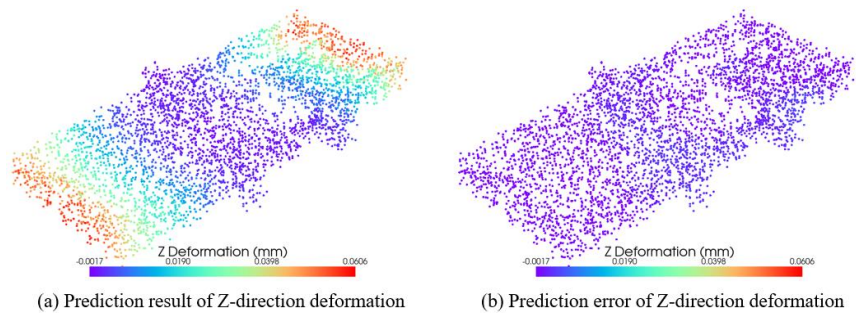

(a) Prediction result of Z-direction deformation  (b) Prediction error of Z-direction deformation

Fig. 11 Prediction result of a frame component example

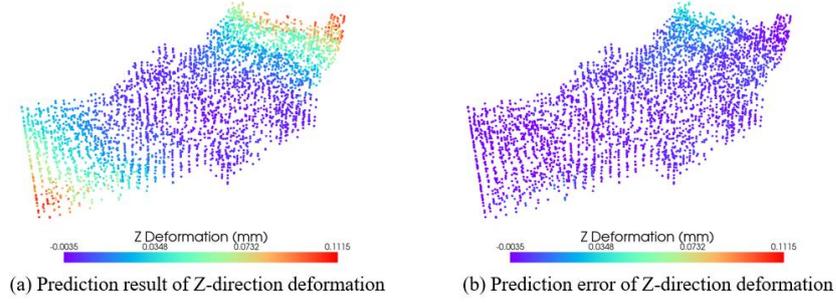

(a) Prediction result of Z-direction deformation  (b) Prediction error of Z-direction deformation

Fig. 12 Prediction result of a C-beam component example

The proposed method is further verified on outputting deformation on all three directions to validate its performance on Multiphysics problem. The result of multi-direction deformation is shown in Table 5 and Table 6. As shown in the verification result, the proposed NDNO outperforms the baseline model Geo-FNO in prediction of both main direction deformation and multi-direction deformation.

Table 5 Multi-direction deformation prediction result of frame components

| - | | NDNO | Geo-FNO |
|---|---|---|---|
| Averaged max error (mm) | | **0.0480** | 0.0901 |
| RMSE (mm) | X | **0.0025** | 0.0029 |
| | Y | **0.0018** | 0.0028 |
| | Z | **0.0141** | 0.0185 |

Table 6 Multi-direction deformation prediction result of C-beam components

| - | | NDNO | Geo-FNO |
|---|---|---|---|
| Averaged max error (mm) | | **0.0283** | 0.0888 |
| RMSE (mm) | X | **0.0015** | 0.0017 |
| | Y | 0.0013 | 0.0013 |
| | Z | **0.0078** | 0.0095 |

As shown in the prediction result on both frame component dataset and C-beam component dataset, the proposed NDNO framework achieved high prediction accuracy in multi-physics field problems. And it outperformed the baseline model in predicting the overall deformation field, especially in the max error in the prediction result. The visualized distribution of the predicted deformation error field also supports that the prediction results globally approached the real deformation field, with po     or max errors in only individual samples. The distribution of max error in NDNO also exhibits to be more clustered at low error zones than the baseline models on tested component types.

### 4.3. Generalization Verification

The proposed NDNO is also tested on component geometry from the same component type that exceeds the range of dimension in training dataset to test its generalization capability on unseen geometry. With frame component geometries ranging from 208~612mm in X direction and 128~248mm in Y direction, the components with x-direction size larger than 400mm and y-direction size larger than 200mm is used for model test. And the remaining components is set as the training dataset.

The model generalization verification result is listed in Table 7 and Table 8, and the distribution of the max errors is shown in Fig. 13. As shown in the result, the NDNO method achieves good result in the unseen component geometries with slight accuracy decrease. And it apparently outperforms the baseline Geo-FNO in this generalization validation. Therefore, NDNO is capable of handling with unseen domain geometries in an identical PDE solving problem with accuracy.

Table 7 Main direction deformation prediction result of component dimension generalization test

| - | NDNO | Geo-FNO |
|---|---|---|
| Max error (mm) | **0.0982** | 0.1336 |
| RMSE (mm) | **0.0301** | 0.0426 |

Table 8 Multi-direction deformation prediction result of component dimension generalization test

| - | | NDNO | Geo-FNO |
|---|---|---|---|
| Averaged max error (mm) | | **0.1116** | 0.1378 |
| RMSE (mm) | X | **0.0044** | 0.0053 |
| | Y | **0.0025** | 0.0026 |
| | Z | **0.0323** | 0.0449 |

Moreover, the generalization capability of NDNO is also tested on unseen component types. A NDNO model trained on dataset of frame components is used to predict the deformation of C-beam components as its test dataset. As shown in Table 9 and Table 10, despite the apparent decrease in accuracy, the model is still capable of predicting the deformation field of component types different from training dataset.

Table 9 Main direction deformation prediction result of component type generalization test

| - | NDNO | Geo-FNO |
|---|---|---|
| Averaged max error (mm) | **0.1102** | 0.1194 |
| RMSE (mm) | **0.0251** | 0.0318 |

Table 10 Multi-direction deformation prediction result of component type generalization test

| - | | NDNO | Geo-FNO |
|---|---|---|---|
| Averaged max error (mm) | | **0.1039** | 0.1140 |
| RMSE (mm) | X | **0.0047** | 0.0055 |
| | Y | **0.0033** | 0.0059 |
| | Z | **0.0260** | 0.0350 |

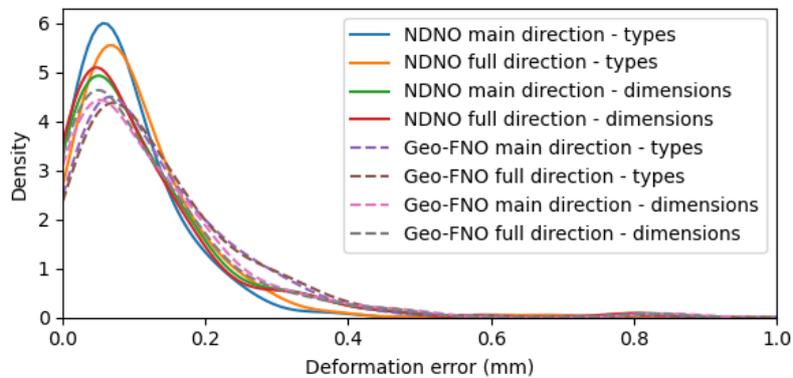

Fig. 13 Density distribution of deformation max error

The proposed NDNO method still outperforms the baseline model in the generalization cases of both component dimensions and types, with the apparent discrepancy in RMSE of predicted deformation field. Though their gap in the averaged max error remains similar, the distribution across all tested parts reveals that the proposed method exhibited higher prediction stability in the generalization test dataset.

*4.4. Ablation Study*

Ablation study was conducted to examine the effect of the proposed diffeomorphic neural network on the prediction accuracy of the NDNO method. The mapped coordinate which is transformed to the reference domain by diffeomorphic neural network is replaced with original coordinate to build the ablated model. Both the NDNO model and the ablation model were trained and tested on the same dataset of frame components and C-beam components. The test result is listed in Table 11-14, including prediction of the main direction deformation and multi-direction deformation.

Table 11 Main direction deformation prediction result of component type generalization test

| - | NDNO | Ablated model |
|---|---|---|
| Max error (mm) | **0.0452** | 0.0528 |
| RMSE (mm) | **0.0140** | 0.0168 |

Table 12 Main direction deformation prediction result of component type generalization test

| - | NDNO | Ablated model |
|---|---|---|
| Max error (mm) | **0.0273** | 0.0300 |
| RMSE (mm) | **0.0076** | 0.0086 |

Table 13 Main direction deformation prediction result of component type generalization test

| - | | NDNO | Ablated model |
|---|---|---|---|
| Averaged max error (mm) | | **0.0480** | 0.0571 |
| RMSE (mm) | X | **0.0025** | 0.0028 |
| | Y | **0.0018** | 0.0020 |
| | Z | **0.0141** | 0.0178 |

Table 14 Main direction deformation prediction result of component type generalization test

| - | | NDNO | Ablated model |
|---|---|---|---|
| Averaged max error (mm) | | **0.0283** | 0.0312 |
| RMSE (mm) | X | **0.0015** | 0.0016 |
| | Y | **0.0013** | 0.0012 |
| | Z | **0.0078** | 0.0089 |

The result of the ablation study reveals that the NDNO method achieved improvement in deformation prediction accuracy by mapping the coordinate data into a common reference through the diffeomorphic neural network. Outperforming the ablation model in metric of deformation prediction on both main direction and multiple directions, NDNO is verified with gains on effectively learning the solution operator of the machining deformation problem in a common reference domain, compared with directly learn in the original domain with identical model structure.

**5. Conclusion**

A NDNO framework is proposed in this paper to learn the solution operator of PDEs governing machining deformation across varying component geometries. This method aims to achieve efficient problem solving in unseen geometries with a single trained model, thereby avoiding the computational cost and time for retraining. Point cloud is adopted as the geometric data representation in this framework as they effectively capture complex shapes and detailed geometric discrepancy. By employing a diffeomorphic neural network that directly learns mappings, the framework is capable of transforming the diverse geometric domains into a common reference domain in an end-to-end manner. Thereby, the neural operator is used to learn the solution operator of the governing PDE in the reference domain. The effectiveness and performance of NDNO were validated through experiments on component machining deformation prediction involving numerous geometries. Furthermore, its generalization capability was verified on dataset containing unseen component dimensions and types, demonstrating its potential for application in problems with varying geometries.

Despite the effectiveness of the proposal, this study still merit further exploration in the following aspects. Firstly, the diffeomorphic neural network learns the transformation between geometries and the reference domain under constraints of similarity and diffeomorphic. Incorporating morphing velocity flow structure into the framework can enhance the diffeomorphism enforcement. Moreover, the size of point cloud data is limited by the embedded transformer modules, which requires sampling method for compression. Future research will explore improving the handling of large-size data for higher accuracy in geometric feature recognition and prediction. Finally, further efforts will be directed towards enhancing the accuracy of output solutions through integration of advanced methods.